\documentclass[journal]{IEEEtran}
\usepackage{stfloats}
\usepackage{graphicx}
\usepackage{caption}
\usepackage{xcolor}
\usepackage{hyperref}
\usepackage{subcaption}
\usepackage[normalem]{ulem}
\usepackage{amsmath}

\ifCLASSINFOpdf

\else

\fi

\begin{document}

\title{Detecting What Matters: A Novel Approach for Out-of-Distribution 3D Object Detection in Autonomous Vehicles}

\author{Menna~Taha,
        Aya~Ahmed,
        Mohammed~Karmoose, 
        and~Yasser~Gadallah,~\IEEEmembership{Senior~Member,~IEEE} 
\thanks{M. Taha is with the Faculty of Engineering at Alexandria University, Alexandria, Egypt (email:\href{mailto:es-mennatahha@alexu.edu.eg} {es-mennatahha@alexu.edu.eg}).}
\thanks{A. Ahmed (email:\href{mailto:aya72@aucegypt.edu} {aya72@aucegypt.edu}) and Y. Gadallah (email:\href{mailto:ygadallah@aucegypt.edu} {ygadallah@aucegypt.edu}) are with the Department of Electronics and Communications Engineering at The American University in Cairo, Egypt.}
\thanks{M. Karmoose is with the Wireless Intelligent Networks Center (WINC), School of Engineering and Applied Sciences (EAS), Nile University, Giza, Egypt, and with the Faculty of Engineering, Alexandria University, Alexandria, Egypt (email: \href{mailto:mkarmoos@nu.edu.eg}{mkarmoos@nu.edu.eg}, \href{mailto:mkarmoose@alexu.edu.eg}{mkarmoose@alexu.edu.eg}).}}

\maketitle

\begin{abstract}
Autonomous vehicles (AVs) use object detection models to recognize their surroundings and make driving decisions accordingly. Conventional object detection approaches classify objects into known classes, which limits the AV's ability to detect and appropriately respond to Out-of-Distribution (OOD) objects. This problem is a significant safety concern since the AV may fail to detect objects or misclassify them, which can potentially lead to hazardous situations such as accidents. Consequently, we propose a novel object detection approach that shifts the emphasis from conventional class-based classification to object harmfulness determination. Instead of object detection by their specific class, our method identifies them as either \textit{harmful} or \textit{harmless} based on whether they pose a danger to the AV. This is done based on the object position relative to the AV and its trajectory. With this metric, our model can effectively detect previously unseen objects to enable the AV to make safer real-time decisions. Our results demonstrate that the proposed model effectively detects OOD objects, evaluates their harmfulness, and classifies them accordingly, thus enhancing the AV decision-making effectiveness in dynamic environments.
\end{abstract}

\begin{IEEEkeywords}
Autonomous vehicles, 3D Object detection, Carla, Out-of-Distribution.
\end{IEEEkeywords}

\IEEEpeerreviewmaketitle

\section{Introduction}

\IEEEPARstart{A}{utonomous} vehicles (AVs), also known as self-driving cars, have the potential to revolutionize transportation by partially or completely replacing the human drivers \cite{Henriksson}. They operate using a variety of sensors, advanced artificial intelligence (AI), including machine learning (ML), algorithms, and other classical solutions to navigate their environment, make decisions, and control operations. As far as the use of ML is concerned, it facilitates data-driven solutions of highly complex data interpretation and control using data-driven approaches. The role of the sensors is to provide data that needs to be processed and interpreted to allow the vehicle to decide how to navigate its environment. One of the most remarkable aspects of AVs is therefore their ability to control operations on their own through the use of advanced artificial intelligence techniques. Their ability to accurately interpret the surrounding environment conditions not only enhances the safety of the vehicle and pedestrians by reducing the potential for human drivers' errors, but also allows for greater efficiency and convenience for the passengers.

\begin{figure}
    \centering
    \includegraphics[width=0.9\columnwidth]{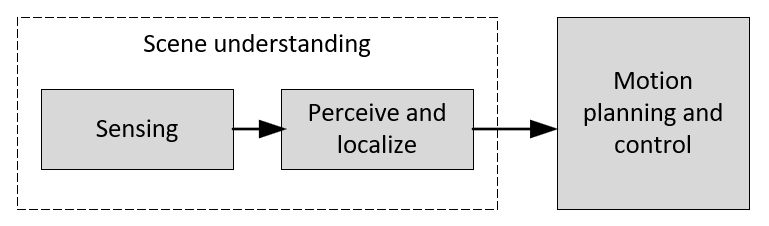}
    \caption{Overview of AV operational architecture}
    \label{fig::AD_arch}
\end{figure}

 Figure \ref{fig::AD_arch} shows an overview of the AV system. The system consists of two main elements of functionality, namely, scene understanding and motion planning and control. One of the key aspects of scene understanding is object detection, which is the ability of the vehicle to identify and locate objects in its vicinity. In many autonomous mobile systems, object detection is one of the most crucial prerequisites to autonomous navigation and safe operation \cite{Huang}. By continuously scanning the environment and making informed decisions based on detected objects, AVs can ensure a safe and smooth journey for their passengers as well as the road pedestrians.

However, a key challenge in object detection in AVs is the identification of Out-of-Distribution (OOD) objects. The most prevalent approach to object detection module development is currently data-driven, where machine learning techniques are used to train such modules to detect and classify objects out of a pre-determined set of classes. The OOD challenge refers to the ability of the object detection module to appropriately react to the presence of objects that do not belong to any of the pre-determined classes. 
OOD detection poses a serious threat to the safety and reliability of AVs \cite{Jingkang}. Although some studies have proposed solutions for the multiclass classification task, the issue of OOD detection in the multilabel classification domain has largely been overlooked and remains underexplored \cite{Wilson}. Existing studies address the challenge of OOD samples by incorporating a separate module that classifies detected objects or samples into OOD or in-distribution (ID) categories.

In this paper, we address the OOD detection challenge in a novel approach where we start by posing the question: what is the important information that a successful object detection module needs to extract from the detected objects? Naturally, an object detection model is designed to identify relevant street-level objects, such as pedestrians, vehicles, and street signs. However, distant objects often do not influence the immediate actions of an AV system. On the other hand, certain objects, such as a deceased animal on the road or an object flying toward the vehicle, are critical for the AV system to detect and react to. These objects, considered OOD, are not included in the pre-defined training set of the object detection module. The module’s response to such objects could result in either missing them or mistakenly detecting OOD objects with overconfident scores as ID category objects. In both scenarios, the AV system could make incorrect decisions based on the corrupted inputs from the module. Such objects may pose a real threat to the AV, as illustrated by the following examples:  

\begin{itemize}
    \item An AV driving inside cities on narrow roads and close-by surrounding buildings, such as within the Gothic Quarter in Barcelona or the Quartieri Spagnoli in Naples, may encounter objects that have fallen from nearby buildings. Such objects can pose an immediate threat and should be detected by the module. However, most of those random objects would potentially be OOD.
    \item A heavy metallic object in the center of the lane could be mistakenly identified as a piece of trash. As a result, the vehicle might continue moving forward, potentially endangering both the vehicle and its driver.
\end{itemize}

The examples above show cases of objects that are considered OOD by conventionally trained object detection modules. We propose to develop a module to detect what essentially matters for the autonomous driving task which is whether an object is classified as harmful or harmless regardless of what it actually is. A {\it harmful} object is something that might potentially harm the AV or its surroundings if the vehicle does not provide an appropriate reaction. In contrast, a {\it harmless} object does not pose an immediate threat to the AV and thus does not cause a notable reaction in the AV. This classification of objects would cover both ID and OOD objects per the conventional classification approach. We show that, by appropriately labeling the objects according to our proposed notion of the harmfulness, our object detection module successfully detects objects that belong to previously unseen classes and correctly labels them as harmful or harmless.\\

\noindent \textbf{Paper Contribution.}
In this paper, we introduce a novel approach to address the problem of OOD object detection. Conventional object detectors react to OOD objects by either ignoring them or falsely classifying them with overconfident scores. However, from a real-life perspective, our system ensures that the vehicle is not exposed to such situations by classifying the objects, regardless of whether they are recognizable, into objects that can be harmful or harmless to the vehicle's operation.
We leverage multi-sensor data and approach the problem from the perspective of 3D object detection. Additionally, to address the issue of having different 3D datasets for training and validation, we use the Carla simulator to generate a custom multi-modal dataset. The primary contributions of this paper are as follows
\begin{itemize}
    \item Introducing a novel approach to address the challenge of OOD objects in AV object detection modules by classifying objects based on the threat they pose to the ego vehicle as opposed to what the object actually is.
    \item Defining a threat-based classification metric to identify harmful and harmless objects in the scene.
    \item Generating a comprehensive dataset featuring various maneuvering scenarios with a range of ID objects, along with a supplementary OOD evaluation dataset to test our proposed solution.
    \item Adapting existing OOD evaluation metrics to effectively measure the performance of the proposed method.
\end{itemize}

The rest of the paper is organized as follows. In Section \ref{sec: related work}, we discuss the related work from the literature. In Section \ref{sec: annotation metric}, the proposed threat-based metric for object classification is discussed in detail. Section \ref{sec: system architecture} discusses the system model and the architecture used to implement our proposed solution. Section \ref{sec: dataset preparation} focuses on the customized dataset generation. The experimental results are then discussed in Section \ref{sec: system evaluation}. In Section \ref{sec: conclusion}, we conclude the study.

\section{Related work} \label{sec: related work}
In this section, we present the related work in OOD object detection, 3D object detection evaluation in AVs, and LiDAR-camera-supported 3D object detection areas of research.
\subsection{OOD Detection} 
The authors in \cite{Henriksson} discuss the importance of OOD detection as machine learning quality assurance methods in the safety life cycle of autonomous driving systems, which improves the system robustness by enabling the model to recognize and respond to unfamiliar input. Although various OOD detection techniques have been developed, most of them focus on image classification, with fewer techniques addressing the object detection tasks such as \cite{Zolfi} and \cite{Wilson}. In \cite{Hendrycks}, a baseline for OOD detection is established by observing that OOD samples tend to have lower softmax probabilities than ID samples, thus allowing images to be classified as ID or OOD. The study in \cite{Liang} improves this baseline by increasing the gap between the softmax scores of the ID and OOD samples. The solutions provided by \cite{Hendrycks} and \cite{Liang} are considered post-hoc methods that can be applied to pre-trained models without the need for retraining \cite{Jingkang}. 
Similarly, the approach in \cite{Wilson} utilizes a post-hoc method to tackle the OOD problem in object detection tasks by extracting sensitive feature vectors from specific layers of the image detector backbone and processing these features through a multilayer perceptron to classify the detected objects as ID or OOD. 

The study in \cite{Huang} extends the OOD detection problem to the context of 3D LiDAR-based object detection. They use a feature extraction method to adapt various OOD detection techniques for 3D object detection. In \cite{Michael}, the authors propose adding a fixed multilayer perceptron (MLP) to the 3D object detector to classify detections as OOD or ID based on features extracted from the feature map of the object detection model. To train the MLP, synthetic OOD objects are generated from ID objects by scaling random ID objects with unusual values. 

\subsection{3D object detection evaluation in autonomous vehicles}
Multiple large-scale publicly available datasets are used to test the performance of object detection methods for autonomous driving. NuScenes \cite{Holgar}, KITTI \cite{kitti}, and Waymo \cite{Waymo} are considered the most widely used datasets due to the wide range of sensor modalities and driving scenarios they provide, which makes them suitable for testing multi-modality object detection methods. 
Another approach for object detection in AVs is generating a custom dataset using the Carla simulator \cite{Carla} due to the myriad sensors and customization methods that the simulator provides. 

To evaluate OOD detection methods, models are trained on an ID dataset and then tested on an OOD dataset from a different distribution, which can be obtained from realistic images \cite{Hendrycks}, \cite{Liang}, \cite{Wilson}, \cite{Michael}  or synthetic noise datasets \cite{Huang}. Currently, there are no multimodal datasets specifically designed to evaluate the performance of the OOD detection methods for 3D object detection. One proposed solution in \cite{Huang} is to use the KITTI dataset and incorporate synthetic OOD objects from other datasets and the Carla simulator during evaluation. However, this approach is unrealistic because the LiDAR point cloud of the OOD objects is concatenated to the original point cloud, which introduces undesired features such as the differences in point cloud intensities around the inserted object. These differences may bias the model, leading to unrepresentative performance results \cite{Michael}. 
Another solution, proposed in \cite{Michael}, involves using existing datasets like NuScenes by treating underrepresented objects in the dataset as OOD objects. 
This method restricts the potential OOD objects to specific classes that are present in the dataset and reduces the variety of ID objects. In this study, Carla is used to generate a customized dataset with various OOD objects and diverse driving scenarios to test the effectiveness of the proposed approach for the OOD objects problem.

\subsection{LiDAR-camera fusion for 3D object detection}
Fusion between LiDAR and camera data can lead to more accurate and reliable object detection results \cite{Mao}. Existing approaches for LiDAR-camera object detection differ according to the method of fusion of the image and LiDAR data. In intermediate-fusion-based methods such as \cite{DeepFusion}, \cite{Futr3d}, \cite{TransFusion} and \cite{BEVFusion}, the LiDAR and image features can be fused at intermediate stages of a LiDAR-based 3D object detector \cite{Mao}. In this work, we adopt BEVFusion \cite{BEVFusion} as our 3D object detector, where image and LiDAR features are fused in a unified bird-eye-view representation at the backbone network of the detector.

\begin{figure*}
    \begin{subfigure}{0.45\linewidth}
        \includegraphics[width=\linewidth]{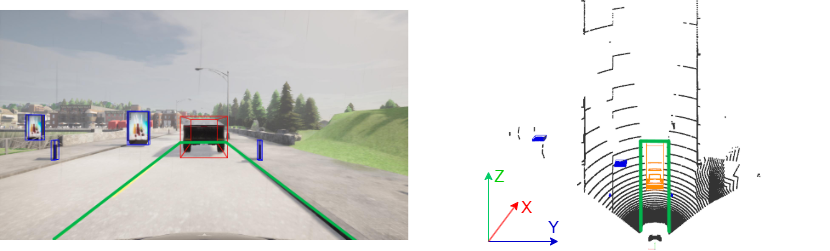}
        \caption{Speed = 5 m/s, steering angle = 0\textdegree.}
        \label{fig:Danger straight}
    \end{subfigure}
    \hfill
    \begin{subfigure}{0.45\linewidth}
        \includegraphics[width=\linewidth]{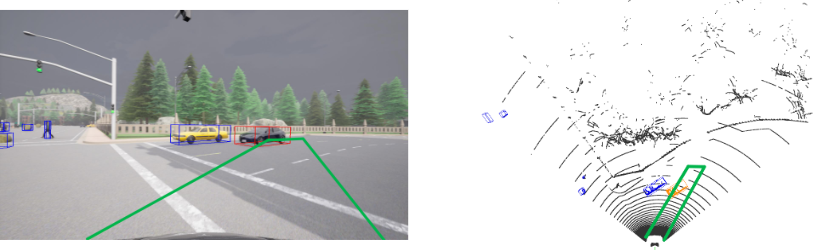}
        \caption{Speed = 8.35 m/s, steering angle = 26.4\textdegree.}
        \label{fig:Danger inclined}
    \end{subfigure}      
    \caption{Danger zone projections in LiDAR and camera views under different vehicle states.}
    \label{fig:Danger zone}
\end{figure*}

\section{Classification Metric} \label{sec: annotation metric}
Our work is based on assessing the environment in terms of harmfulness to the ego vehicle. The harm an object imposes on the ego vehicle depends on several factors, including the nature of the object, its location relative to the car, and the expected vehicle trajectory. Consequently, defining a metric to assess the harmfulness of the object within the scene is not straightforward.

As an initial step in this research, we adopt a simplified metric that considers the object location relative to the vehicle and the ego vehicle speed and steering angle to estimate its trajectory in the immediate future. Specifically, we establish a virtual {\it danger zone} ahead of the car to categorize detected objects into two groups, where those within the zone are deemed harmful, while the others outside the zone are considered harmless. The coordinates and dimensions of the danger zone are determined according to the speed of the car and the steering angle in each frame. We adopt this metric as a proof-of-concept for harmful/harmless object detection. Figure \ref{fig:Danger zone} shows an example of the danger zone in LiDAR and camera frames. Objects inside the green-bordered zone are bounded by red boxes, indicating that they are labeled as harmful, while blue boxes outside the danger zone indicate harmless objects.

\subsection{Danger zone representation}
The zone borders are represented by two forward 3D vectors ($x$, $y$, $z$). The $z$ coordinates, which indicate the depth of the zone, vary depending on the speed of the ego vehicle, while the $x$ and $y$ coordinates, representing the orientation of the vectors relative to the car, change according to the steering angle. The zone is enclosed by a third vector normal to the two forward vectors. The zone width, which corresponds to the separation between the two forward vectors, is determined experimentally according to the width of the lane.

The depth of the danger zone varies with the vehicle's speed where higher speeds result in a deeper zone. When the ego vehicle is not moving, the danger zone depth is set to a predefined minimum safe distance between the car and the other objects. The linear relationship between the car's speed and the zone depth is defined in equation \ref{eq:danger zone depth}. The choice of parameters is based on extensive testing to determine a suitable depth that ensures objects fall within the danger zone before the car stops prematurely.
\begin{equation} \label{eq:danger zone depth}
    Depth = speed (m/s) * 2 + min\_safe\_distance
\end{equation}

The inclination of the created zone varies with the steering angle, which determines the ego vehicle's potential path. This inclination is calculated by rotating the forward vector of the vehicle's path around the $z$-axis by an angle equal to the vehicle's steering angle as described by equation \ref{eq:danger zone rotation}.

\begin{equation} \label{eq:danger zone rotation}
    \begin{bmatrix}
    x \\
    y \\
    z 
    \end{bmatrix} = 
    \begin{bmatrix}
    cos\theta & -sin\theta & 0\\
    sin\theta & cos\theta & 0\\
    0 & 0 & 1
    \end{bmatrix} * 
    \begin{bmatrix}
    x\_forward \\
    y\_forward \\
    z\_forward 
    \end{bmatrix}
\end{equation}
where $\theta$ represents the ego vehicle's steering angle. Figure \ref{fig:Danger inclined} illustrates how the vectors defining the danger zone rotate in accordance with the steering angle.

The danger zone appears as a trapezoid in front camera images and as a rectangle moving with the vehicle in the LiDAR bird-eye view point cloud, as shown in Figure \ref{fig:Danger zone}.  

\subsection{Object category}
The decision to classify an object as harmful depends on the overlap between the object base and the danger zone, where an object is annotated as harmful if this overlap exceeds a specified threshold area or a certain ratio of the base total dimensions. For instance, in Figure \ref{fig:Danger straight}, the car is entirely within the danger zone, so it is marked as harmful. Similarly, the car in Figure \ref{fig:Danger inclined} is also marked as harmful, even though it is only partially inside the danger zone, because the portion within the zone exceeds the defined thresholds.

\begin{figure*}
    \centering
        \centering
        \includegraphics[width=0.9\linewidth]{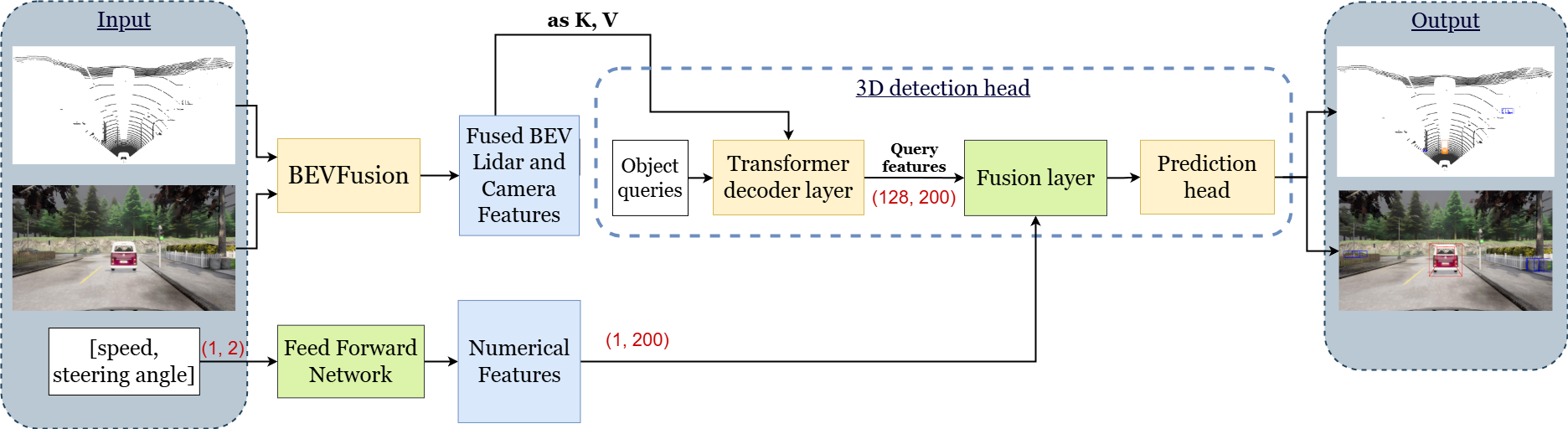}
        \caption{Model architecture highlighting shared layers with BEVFusion (orange) and added numerical feature processing layers (green). Arrow labels indicate vector dimensionality at each stage. Numbers on the arrows represent the vectors dimensionality in each stage.}
        \label{fig:Model architecture}
\end{figure*}
\section{System Architecture} \label{sec: system architecture}
\subsection{Overview} \label{sec: overview}
Our approach differs from existing solutions as follows. While most systems that address the OOD problem focus on categorizing scene objects as either ID or OOD, our system differs by using a specific metric to classify objects from any distribution based on their potential risk. We transform the problem into a binary object detection task as the objects are labeled as harmful or harmless. Our solution addresses the object detection task and the OOD problem simultaneously by training the object detector to locate objects from any distribution and classify them as harmful or harmless. Each object is annotated according to the metric detailed in Section \ref{sec: annotation metric} and is then utilized for model training and testing. The calculated danger zone coordinates are used to label the objects before model training. During both the training and the inference phases, the model has no access to the actual danger zone coordinates but rather it receives numerical data such as the speed and the steering angle that were used to establish this zone. 
In essence, the model can learn the labeling metric using the provided information for object classification without having prior knowledge of the metric specification.

It is important to note that the vehicle state data, which includes the speed and steering angle, is essential because the same frame can yield different object labels based on these variables. For example, a car with a specific state, i.e., a specific speed and steering angle, classifies objects differently than a car with a different state due to the different expected trajectories of the vehicle in the immediate future.

\subsection{Model Architecture}
We consider a vehicle equipped with a 3D LiDAR sensor and an RGB camera. The 3D LiDAR-camera-based object detector is fed with the output of these sensors in the form of a front-camera image and a LiDAR point cloud. 
In addition, the detector receives the vehicle speed and steering angle as numerical inputs. The model output consists of a set of detected objects, each annotated with a 3D bounding box and classified as either harmful or harmless.

For the 3D object detection task, we adopt a modified BEVFusion model architecture. As illustrated in Figure \ref{fig:Model architecture}, our model architecture comprises two main branches. The first branch uses the BEVFusion model, which integrates the LiDAR and camera data in a unified bird's-eye-view representation before passing it through the transformer decoder layer in the 3D object detection head. The second branch processes numerical data, with the vehicle speed and steering angle passing through a feed-forward network for numerical feature extraction. A convolution-based fusion is applied to integrate the numerical features of dimensions (1, 200) with the decoder output of dimensions (128, 200). The combined camera, LiDAR, and numerical features are then processed through the prediction head for object localization and classification. The 3D object detection head architecture is based on the \cite{BEVFusion} and \cite{TransFusion} approaches, employing a class-specific center heatmap head to localize the objects' centers and regression heads to predict the dimensions and rotation of the bounding boxes.

% Changes to model configurations
By default, the BEVFusion model takes images from six cameras and a LiDAR point cloud covering a 360\textdegree{} horizontal field of view. However, according to our metric, harmful objects appear only in the front camera, and therefore objects in the other five cameras are always harmless. This leads to a significant class imbalance in our dataset. To address this, we modified the model to use only the front camera and limited the LiDAR field of view to match the front camera's field of view. This aligns with how experienced human drivers consider objects as harmful or harmless based on their view of objects through the front window. The detector is trained on an ID dataset that includes a known set of objects to localize and classify as harmful or harmless. During the evaluation process, a test dataset is used where each frame contains OOD objects alongside ID objects, as detailed in Section \ref{sec: dataset preparation}.

\section{Dataset Preparation} \label{sec: dataset preparation}
\subsection{Dataset Selection}
Our dataset selection process is based on two main criteria. First, we aim to maintain a reasonable class balance between the harmful and harmless classes to avoid biases in the model results. Second, we need to use an OOD testing dataset that comes from a distribution different from the training dataset while maintaining consistent sensor configurations.
Initially, we considered using the NuScenes dataset due to its popularity as a large-scale autonomous driving dataset that features a LiDAR and six cameras. However, the NuScenes dataset proved unsuitable for our task for two main reasons

\begin{enumerate}
    \item 
    The LiDAR point cloud covers 360\textdegree{} around the ego vehicle. According to our classification metric, all objects around the car were considered harmless, except for some objects in the front camera field of view, which were deemed harmful. Consequently, re-annotating the entire dataset based on this metric, introduced a significant class imbalance%between the harmful and harmless classes
    , which biased the model results towards the harmless class.

    \item 
    Having an OOD evaluation dataset is relatively straightforward for 2D models, where multiple image datasets can be combined and used together. However, for 3D models, autonomous driving datasets vary in sensor configurations and setups. As a result, a model trained on one dataset is not suitable when testing on another dataset. Consequently, we could not use another dataset for OOD evaluation for a model trained on NuScenes.
\end{enumerate}

To address these challenges, we create the dataset using the Carla simulator \cite{Carla}, an open-source simulator for testing and training autonomous driving models. To address the class imbalance issue, we limit the LiDAR sensor field of view to the area within the front camera field of view, as this is where harmful objects can be found. The sensors setup and procedures for generating the ID training dataset and OOD evaluation dataset are identical, with the only difference being the introduction of new OOD objects in the evaluation dataset that are not present in the training dataset. The ID training dataset is used for model training and validation, while the OOD evaluation dataset is only used for testing.

\subsection{Dataset Generation using Carla}
\subsubsection{Ego vehicle setup} 
\begin{figure}
    \centering
    \includegraphics[width=0.9\columnwidth]{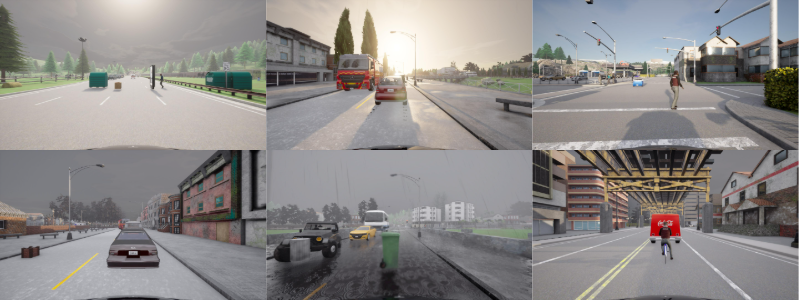}
    \caption{Dataset samples}
    \label{fig:dataset_samples}
\end{figure}

The ego vehicle is equipped with an RGB camera sensor and a LiDAR ray-cast sensor. It autonomously navigates the map, collecting sample data every 0.3 seconds. Each data instance includes an RGB image, a LiDAR point cloud, object annotations, bounding boxes, vehicle speed and steering angle, along with the geometric transformation matrices between different sensor frames.
The data collected from the simulator is then saved in a format compatible with the model configuration. To increase the variations in our dataset, we gather data from different Carla town maps under various weather conditions. We also enhance the diversity of objects on the map by adding static props alongside normal cars, pedestrians, and environmental objects. Figure \ref{fig:dataset_samples} shows samples of the dataset, including different objects in various towns and weather conditions.

\subsubsection{Data labeling} 
We restrict the saved instances to objects appearing in the front camera field of view. For object classification, our metric is applied to define the danger zone in the map. Objects are classified into two classes, namely, harmful and harmless, depending on the object's position relative to this danger zone.
 
 \subsubsection{ID training Data statistics}
 We collected a total of around 43000 samples including 33802 samples for training and 9066 samples for validation. The objects present in our dataset are summarized in Table \ref{table: ID Dataset}. 
 
\begin{table}[t]
\centering
    \caption{The ID Dataset Statistics}
    \label{table: ID Dataset}
    \begin{tabular}{|c|c c|c|c c|}
     \hline
     \multicolumn{3}{|c|}{Train}                  & \multicolumn{3}{|c|}{Validation} \\
        \hline Vehicles       & harmful  & 9146   & Vehicles       & harmful  & 2578 \\
                              & harmless & 76070  &                & harmless & 18054 \\
        \hline Pedestrians    & harmful  & 940    & Pedestrians    & harmful  & 172 \\
                              & harmless & 16280  &                & harmless & 3164 \\
        \hline Static objects & harmful  & 2535   & Static objects & harmful  & 496 \\
                              & harmless & 121632 &                & harmless & 26063 \\
        \hline Total          & harmful  & 12621  & Total          & harmful  & 3246 \\
                              & harmless & 213982 &                & harmless & 47281 \\
          \hline
    \end{tabular}
    
\end{table}

\subsubsection{OOD evaluation dataset statistics}
For the system evaluation, we generate a test dataset comprising objects similar to those in the training and validation datasets, as well as OOD objects that the model had not encountered during training. This approach allows us to assess the model's ability to detect OOD objects according to the specified metric. We introduce 11 OOD objects in Carla: a barrel, mailbox, construction cone, container, cloth container, advertisement, hay bale, shopping bag, map table, cardboard box, and newspaper box. Table \ref{table: OOD Dataset} presents the statistics of the generated OOD dataset distributed among 6351 data samples. The ratio between OOD objects and ID objects is approximately 0.4. Additionally, the OOD objects are labeled as harmful and harmless, with ratios that closely mirror the categories in the training and validation datasets.

\begin{table}[t]
\centering
    \caption{The OOD Dataset Statistics}
    \label{table: OOD Dataset}
    \begin{tabular}{|c|c c|}
        \hline Vehicles       & harmful       & 1560  \\
                              & harmless      & 15343 \\
        \hline Pedestrians    & harmful       & 402   \\
                              & harmless      & 4676  \\
        \hline Static objects & harmful: ID   & 63    \\
                              & harmless: ID  & 21251 \\
                              & harmful: OOD  & 1043  \\
                              & harmless: OOD & 16268 \\
        \hline Total          & harmful: ID   & 2025  \\
                              & harmless: ID  & 41270 \\
                              & harmful: OOD  & 1043  \\
                              & harmless: OOD & 16268 \\
         \hline  
    \end{tabular}
\end{table}

\section{System Evaluation} \label{sec: system evaluation}
    \subsection{Model Training} 
\begin{figure}
    \centering
    \includegraphics[width=0.9\columnwidth]{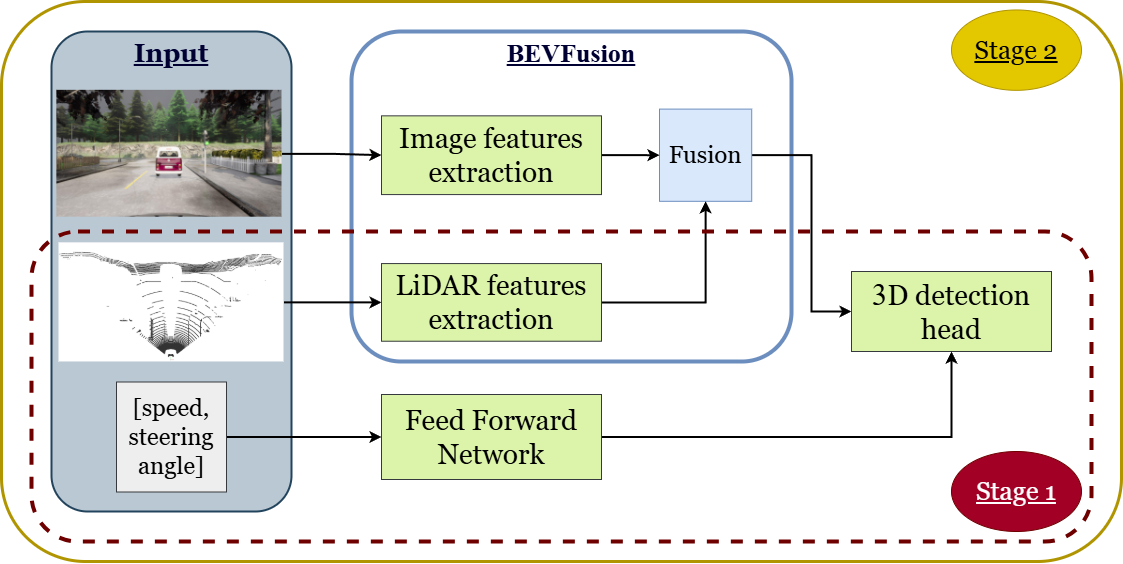}
    \caption{Model training process.}
    \label{fig:training setup}
\end{figure}

    We implemented the model architecture in PyTorch using the MMDetection3D framework \cite{mmdet3d}. The BEVFusion backbone architecture is used for LiDAR and image feature extraction, and a branch for numeric data extraction is added. 
    The image and non-image data are then fused before being passed through the prediction head for object localization and classification. A pre-trained Swin\_T model \cite{swinT} is used as the image backbone, while the rest of the architecture was retrained from scratch. During training, we remove the data augmentation preprocessing step, as the current augmentation techniques do not modify the numerical data according to the updated image and point cloud. 
    As shown in Figure \ref{fig:training setup}, the training process is divided into two stages, where we initially train the LiDAR backbone and the numerical data branch alongside the object detection transfusion head for 10 epochs. The integrated camera and LiDAR data are then used for an additional 3 training epochs. The model is trained using the ID dataset mentioned in Section \ref{sec: dataset preparation}, and the OOD dataset is used for model evaluation.

    \subsection{Results}
    \subsubsection{Comparison to other solutions.}
    The primary distinction of our method from existing approaches is that our approach aims to \textbf{detect} objects and then \textbf{classify} them as harmful or harmless. In contrast, the prevalent approach for OOD detection in the literature assumes an existing object classification module, followed by an OOD detection module that determines whether the classification outputs correspond to ID or OOD objects. 
    Our approach is therefore considered as detection-based, whereas prior methods are classification-based. Consequently, standard OOD evaluation metrics such as the Area Under the Receiver Operating Characteristic (AUROC) and the Area Under the Precision-Recall (AUPR) curve are not directly applicable in our approach. \cite{Peng}.
    
    \subsubsection{Model evaluation on the ID dataset}  
    To evaluate the model's performance on the ID dataset, we utilize the mean Average Precision (mAP) metric by following the NuScenes evaluation approach \cite{Holgar}, where a prediction is matched to a ground truth based on the distance between the centers of the prediction box and the ground truth box. The average precision (AP) is calculated using the recall versus precision curve for different distance matching thresholds. These values are then averaged to obtain the mAP. 
    
    \subsubsection{Model evaluation on the OOD dataset} 
    To evaluate the OOD detection performance, we use an OOD test dataset comprising both the OOD objects, which were not encountered during training, and the ID objects. We apply three variations of the mAP as our evaluation metrics. First, we calculate the mAP for the entire OOD dataset to compare the model's performance before and after introducing the OOD objects. Given the smaller number of the OOD objects in the dataset compared to the ID objects, as typically seen in real-world scenarios, we chose to split the mAP calculation, once for the ID objects and once for the OOD objects only. In this variation, we separate true positives (TPs), which are predictions that are correctly matched to a ground truth, into ID TPs and OOD TPs for recall and precision calculations. We also separate false positives (FPs), which are predictions matched to a ground truth of a different class or unmatched to any ground truth object. If an object is correctly detected but misclassified, it is added as an FP to the distribution of the misclassified object. However, if an FP object is detected but not matched to any ground truth bounding box, it is considered an OOD. For further elaboration, we calculate the separated mAP with a third variation, where unmatched FPs are ignored and only the misclassified objects are considered as FPs. 

\begin{figure}
        \centering
        \includegraphics[width=0.9\columnwidth]{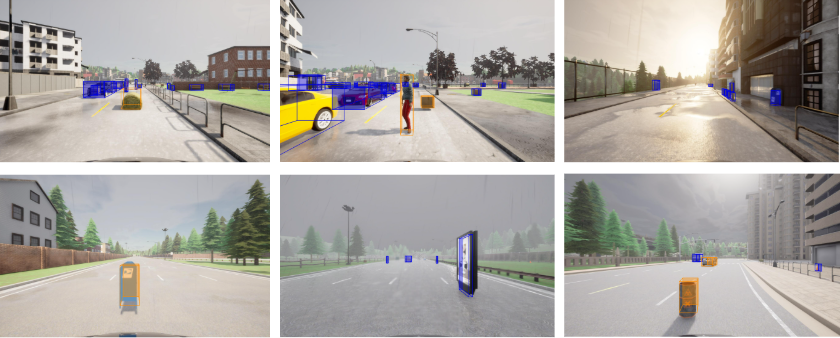}
        \caption{Sample outputs for OOD object detection}
        \label{fig:reults_samples}
    \end{figure}

    Table \ref{table: OOD model evaluation} shows the model's results on the validation dataset, which contains only ID objects, as well as the evaluation results on the dataset containing both ID and OOD objects. The evaluation results display the separated mAP for ID and OOD objects in two different variations, along with the combined mAP for the entire evaluation dataset. The results indicate that the combined mAP for ID and OOD objects is affected by the introduction of OOD objects. For the harmless class, the mAP on the validation dataset is 0.83, while it decreased to 0.72 on the OOD evaluation dataset. For harmful objects, the mAP on the validation dataset is 0.79, while it goes down to 0.715 on the OOD evaluation dataset. The second metric results, which consider all unmatched FPs as OOD objects, demonstrate good performance for harmful objects where the mAP reaches 0.653, despite the penalization caused by assuming all unmatched background objects as OOD. Conversely, the mAP for harmless objects significantly decreases to 0.344, which is due to the large number of background objects being classified as harmless compared to the harmless OOD objects, while fewer background objects are classified as harmful. This behavior aligns with the behavior of an experienced human driver, where a driver typically assumes any distant, unknown structure is harmless until it is approached and evaluated for potential harm. This justification is reinforced via the third metric, which shows that when unmatched FPs are excluded, the mAP for harmless objects notably increases from 0.344 to 0.722, while the mAP for harmful objects slightly increases by 0.06. Figure \ref{fig:reults_samples} shows the detection results for various OOD objects, where some of them belong to the harmful category and others are harmless.

    \begin{table*}[t]
    \centering
     \caption{Model Evaluation on the ID and OOD Datasets}
     \label{table: OOD model evaluation}
        \begin{tabular}{|c|c||c|c|c|c|}
            \hline
            &                Training   & \multicolumn{4}{|c|}{Evaluation} \\
            &                ID dataset & \multicolumn{4}{|c|}{OOD dataset} \\
            \hline
             Object category &mAP       &mAP       &mAP\_ID            & mAP\_OOD  & mAP\_OOD \\
                             &(total)   &(total)   &(matched FPs only) & (All FPs) & (matched FPs only) \\
             \hline
              harmful        &0.7947    &0.715     &0.74              & 0.652     & 0.714   \\
              \hline 
              harmless       &0.8289    &0.723     &0.88               & 0.345     & 0.725   \\
              \hline
              Total          &0.8118    &0.719     &0.81               & 0.499     & 0.72   \\
             \hline
        \end{tabular}
    \end{table*}

\begin{figure*}[!t]
    %\centering
    \begin{subfigure}{0.45\linewidth}
        \includegraphics[width=\linewidth]{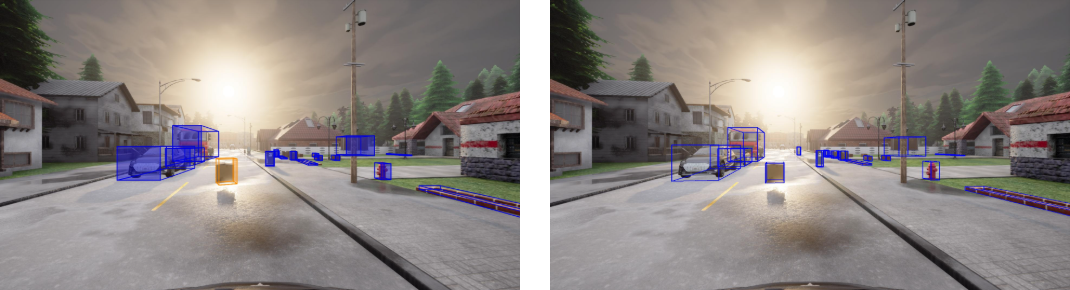}
        \caption{Harmless object falsely detected as harmful; left: prediction, right: ground truth.}
        \label{fig:False harmful}
    \end{subfigure}
    \hfill
    \begin{subfigure}{0.45\linewidth}
        \includegraphics[width=\linewidth]{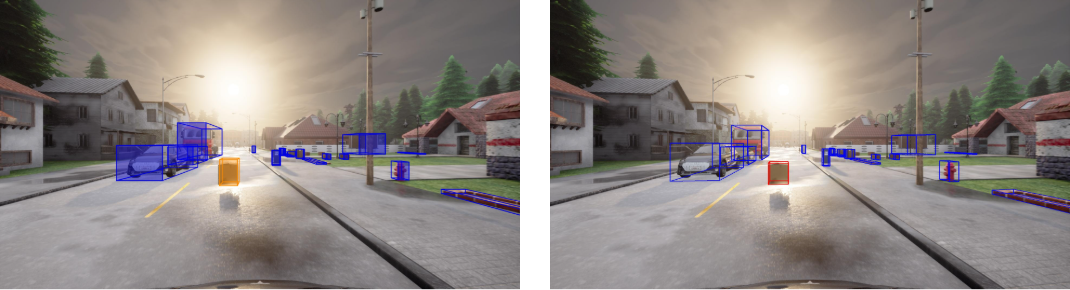}
        \caption{Same object labeled and detected as harmful in next frame; left: prediction, right: ground truth.}
        \label{fig:True harmful}
    \end{subfigure}      
    \caption{Example of early harmful object misclassification.}
    \label{fig:Misclassifications}
\end{figure*}

    An analysis of the results shows that the most frequent misclassification occurs when harmless objects are identified as harmful one or two frames earlier than they are actually harmful. Figure \ref{fig:Misclassifications} shows an example of a box predicted by the model to be harmful, although it is harmless in the ground truth dataset. However, in the next saved frame, the box becomes harmful and the model correctly identifies it.
    
\subsection{Effect of Speed and Steering Angle}
Numerical data (ego vehicle speed and steering angle) are essential for the model to follow the specified metric.

\subsubsection{Ego vehicle speed}
\begin{figure}
    \centering
    \begin{subfigure}{0.48\linewidth}
        \includegraphics[width=\linewidth]{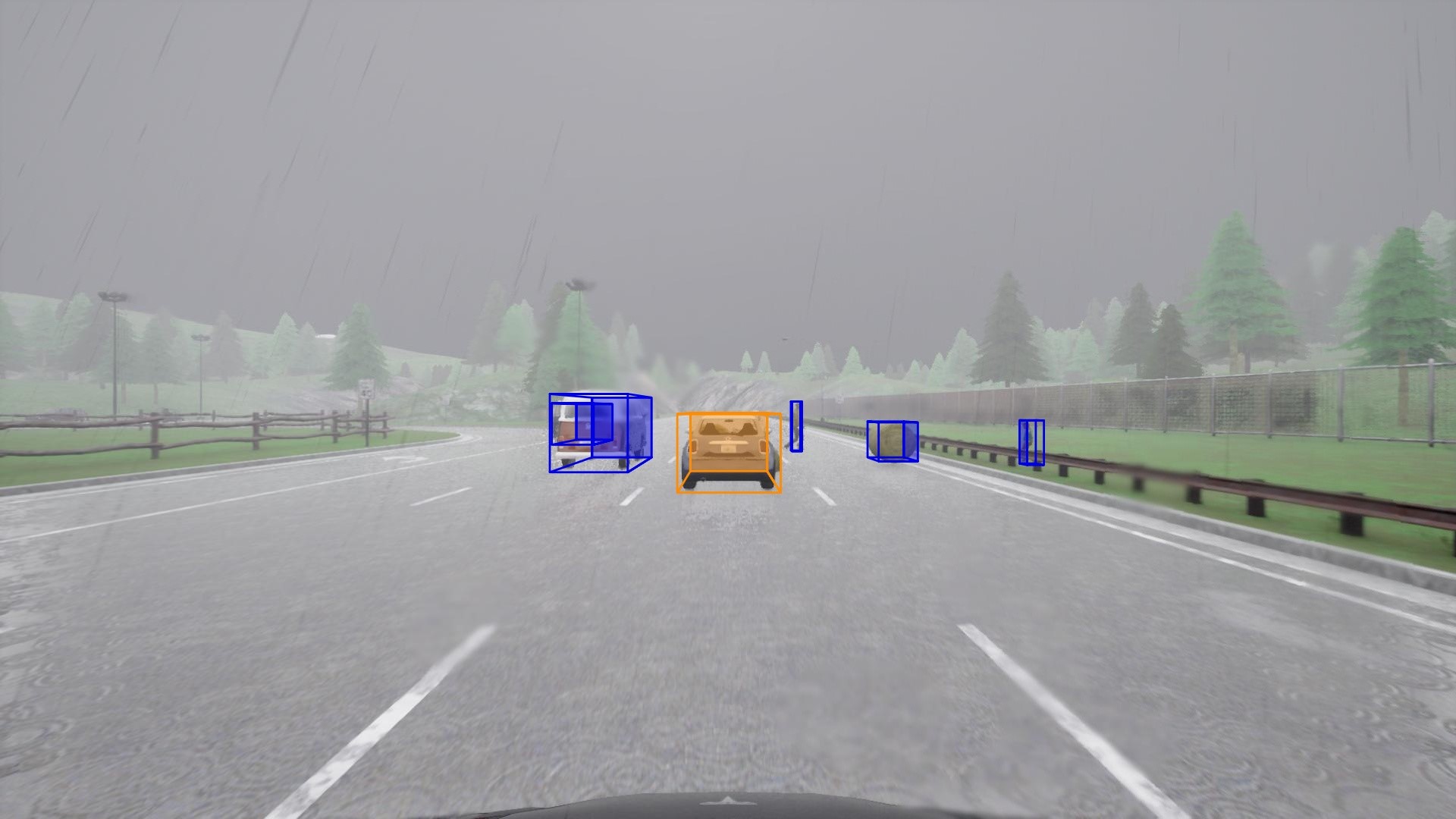}
        \caption{Object detected as harmful (speed = 6.6 m/s).\\}
        \label{fig:speed 6.6}
    \end{subfigure}
    \hfill
    \begin{subfigure}{0.48\linewidth}
        \includegraphics[width=\linewidth]{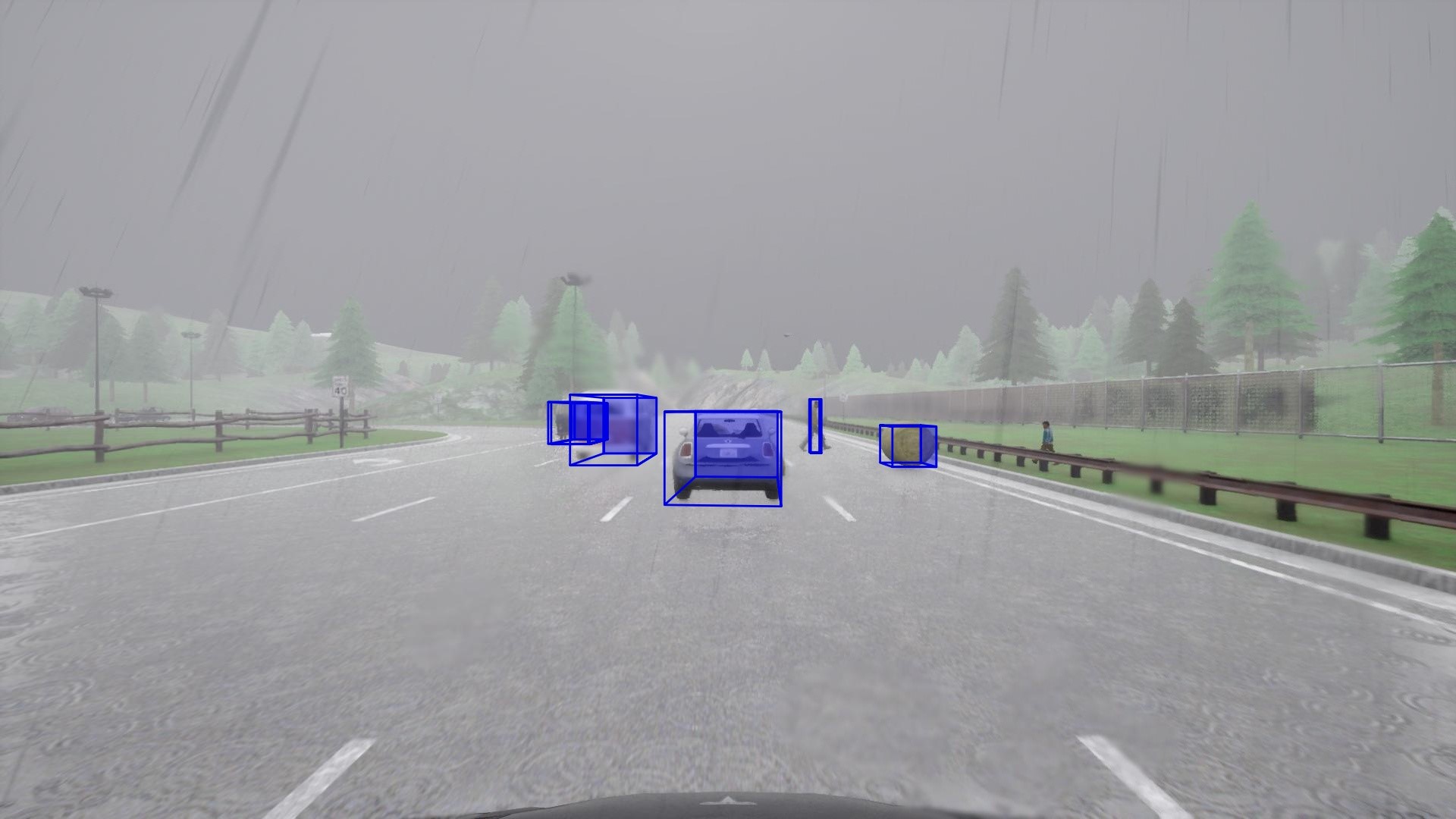}
        \caption{Same object detected as harmless although it is closer to the ego vehicle (speed = 1.4 m/s).}
        \label{fig:speed 1.4}
    \end{subfigure}      
    \caption{Effect of ego vehicle speed on object classification.}
    \label{fig:Speed effect}
\end{figure}

According to our metric, higher speeds result in more distant objects being classified as harmful, where the model successfully learns this criterion, as illustrated in Figure \ref{fig:Speed effect}, which demonstrates how varying speeds lead to different data classifications.

\subsubsection{Ego vehicle steering angle}
\begin{figure}[t]
    \centering
    \begin{subfigure}{0.48\linewidth}
        \includegraphics[width=\linewidth]{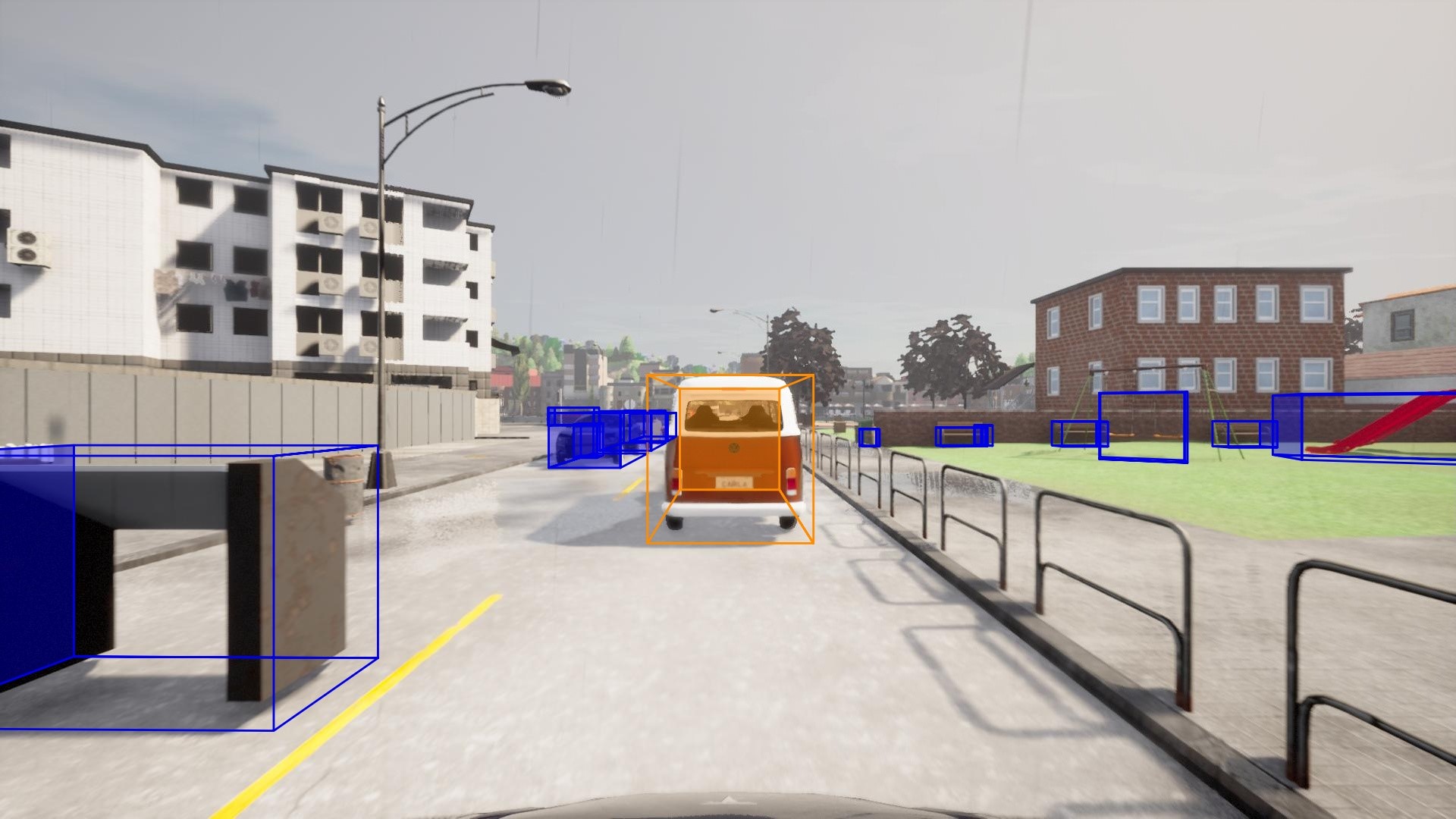}
        \caption{Object in the front detected as harmful (steering angle = 0\textdegree).}
        \label{fig:steer 0}
    \end{subfigure}
    \hfill
    \begin{subfigure}{0.48\linewidth}
        \includegraphics[width=\linewidth]{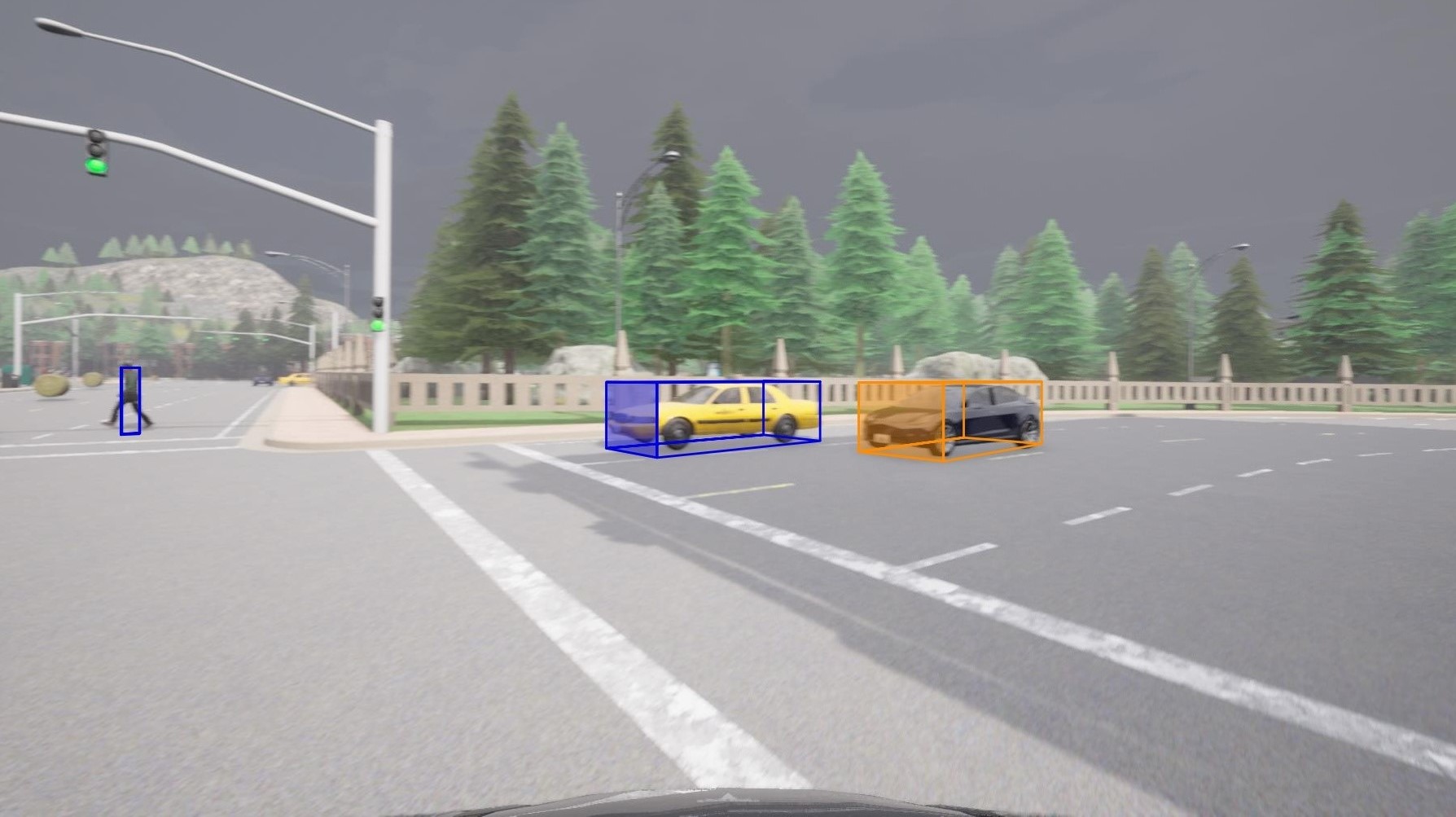}
        \caption{Object on the right detected as harmful (steering angle = 26\textdegree).}
        \label{fig:steer 26 degrees}
    \end{subfigure}      
    \caption{Effect of ego vehicle steering angle on object classification.}
    \label{fig:Steering effect}
\end{figure}

The ego vehicle's steering angle indicates the vehicle's trajectory in the immediate future. Accordingly, the danger zone is inclined by this angle. Figure \ref{fig:Steering effect} shows how the trained object detector reacts to different steering angles. In case of a zero steering angle, only objects inside the ego vehicle's lane can be assumed to be harmful. However, in the case of steering, objects to the right/left of the car can be considered harmful according to the steering direction.

\section{Conclusion} \label{sec: conclusion}
In this paper, we addressed the issue of OOD object detection for AVs by introducing a novel approach that shifts the emphasis from conventional object classification to the harmfulness determination of the detected objects. Contrary to conventional methods, our classification approach uses a model that labels the objects as simply \textit{harmful} or \textit{harmless} without the need to determine their true identity. By training an object detection model using this metric, we demonstrated how AVs can react more effectively to unfamiliar objects and make logical decisions in the interest of road safety. Unlike conventional methods that struggle with unseen classes, our approach consistently applies the harmfulness estimation metric across various object distributions, thus reducing the misclassification errors and improving the AV's decision-making. Our findings highlight the necessity of re-evaluating conventional object detection methods in AVs and shifting toward risk-based object classification to make the AVs safer, more reliable, and more flexible. In future research, we plan to investigate more sophisticated harmfulness metrics design and explore the integration of risk-based object detection modules with other AV perception modules to fine-tune the determination of the harmfulness state of detected objects.

\ifCLASSOPTIONcaptionsoff
  \newpage
\fi


\begin{thebibliography}{1}
\bibitem{Henriksson}
J. Henriksson et al., “Out-of-distribution detection as support for autonomous driving safety lifecycle,” in Lecture Notes in Computer Science, vol. 13975, pp. 233–242, 2023, doi: 10.1007/978-3-031-29786-1\_16.

\bibitem{Huang}
C. Huang et al., “Out-of-distribution detection for LiDAR-based 3D object detection,” in Proc. IEEE Int. Conf. Intell. Transp. Syst. (ITSC), pp. 4265–4271, Oct. 2022, doi: 10.1109/ITSC55140.2022.9922026.

\bibitem{Hendrycks}
D. Hendrycks and K. Gimpel, “A baseline for detecting misclassified and out-of-distribution examples in neural networks,” in Proc. Int. Conf. Learn. Represent. (ICLR), 2017.

\bibitem{Liang}
S. Liang, Y. Li, and R. Srikant, “Enhancing the reliability of out-of-distribution image detection in neural networks,” in Proc. Int. Conf. Learn. Represent. (ICLR), 2017.

\bibitem{Wilson}
S. Wilson, T. Fischer, F. Dayoub, D. Miller, and N. Sünderhauf, “SAFE: Sensitivity-aware features for out-of-distribution object detection,” in Proc. IEEE/CVF Int. Conf. Comput. Vis. (ICCV), pp. 23565–23576, Oct. 2023.

\bibitem{Zolfi}
A. Zolfi et al., “YOLOOD: Utilizing object detection concepts for multi-label out-of-distribution detection,” in Proc. IEEE/CVF Conf. Comput. Vis. Pattern Recognit. (CVPR), vol. 34, pp. 5788–5797, Jun. 2024, doi: 10.1109/CVPR52733.2024.00553.

\bibitem{BEVFusion}
Z. Liu et al., “BEVFusion: Multi-task multi-sensor fusion with unified bird’s-eye view representation,” in Proc. IEEE Int. Conf. Robot. Autom. (ICRA), May 2023, doi: 10.1109/ICRA48891.2023.10160968.

\bibitem{DeepFusion}
Y. Li et al., “DeepFusion: LiDAR-camera deep fusion for multi-modal 3D object detection,” in Proc. IEEE/CVF Conf. Comput. Vis. Pattern Recognit. (CVPR), pp. 17161–17170, Jun. 2022, doi: 10.1109/CVPR52688.2022.01667.

\bibitem{Futr3d}
X. Chen, T. Zhang, Y. Wang, Y. Wang, and H. Zhao, “FUTR3D: A unified sensor fusion framework for 3D detection,” in Proc. IEEE/CVF Conf. Comput. Vis. Pattern Recognit. Workshops (CVPRW), pp. 172–181, Jun. 2023, doi: 10.1109/CVPRW59228.2023.00022.


\bibitem{TransFusion}
X. Bai et al., “TransFusion: Robust LiDAR-camera fusion for 3D object detection with transformers,” in Proc. IEEE/CVF Conf. Comput. Vis. Pattern Recognit. (CVPR), pp. 1080–1089, Jun. 2022, doi: 10.1109/CVPR52688.2022.00116.

\bibitem{kitti}
A. Geiger, P. Lenz, and R. Urtasun, “Are we ready for autonomous driving? The KITTI vision benchmark suite,” in Proc. IEEE Conf. Comput. Vis. Pattern Recognit. (CVPR), pp. 3354–3361, Jun. 2012, doi: 10.1109/CVPR.2012.6248074.

\bibitem{Holgar}
H. Caesar et al., “nuScenes: A multi-modal dataset for autonomous driving,” in Proc. IEEE Conf. Comput. Vis. Pattern Recognit. (CVPR), pp. 11618–11628, Jun. 2020.

\bibitem{Waymo}
P. Sun et al., “Scalability in perception for autonomous driving: Waymo open dataset,” in Proc. IEEE/CVF Conf. Comput. Vis. Pattern Recognit. (CVPR), pp. 2443–2451, Jun. 2020, doi: 10.1109/CVPR42600.2020.00252.

\bibitem{Carla}
A. Dosovitskiy, G. Ros, F. Codevilla, A. M. López, and V. Koltun, “CARLA: an open urban driving simulator,” Proc. Conf. Robot Learn., pp. 1–16, Oct. 2017, [Online]. Available: http://proceedings.mlr.press/v78/dosovitskiy17a/dosovitskiy17a.pdf

\bibitem{Michael}
M. Kösel, M. Schreiber, M. Ulrich, C. Gläser, and K. Dietmayer, “Revisiting out-of-distribution detection in LiDAR-based 3D object detection,” in Proc. IEEE Intell. Veh. Symp. (IV), pp. 2806–2813, Jun. 2024, doi: 10.1109/IV55156.2024.10588849.

\bibitem{mmdet3d}
MMDetection3D Contributors, "MMDetection3D: OpenMMLab next-generation platform for general 3D object detection," 2020. [Online]. Available: https://github.com/open-mmlab/mmdetection3d. 

\bibitem{Jingkang}
J. Yang, K. Zhou, Y. Li, and Z. Liu, “Generalized out-of-distribution detection: A survey,” Int. J. Comput. Vis., vol. 132, no. 12, pp. 5635–5662, Jun. 2024, doi: 10.1007/s11263-024-02117-4.

\bibitem{Mao}
J. Mao, S. Shi, X. Wang, and H. Li, “3D object detection for autonomous driving: A comprehensive survey,” Int. J. Comput. Vis., vol. 131, no. 8, pp. 1909–1963, Apr. 2023, doi: 10.1007/s11263-023-01790-1.

\bibitem{Peng}
P. Cui and J. Wang, “Out-of-distribution (OOD) detection based on deep learning: A review,” Electronics, vol. 11, no. 21, p. 3500, 2022, doi: 10.3390/electronics11213500.

\bibitem{swinT}
Z. Liu et al., “Swin transformer: Hierarchical vision transformer using shifted windows,” in Proc. IEEE/CVF Int. Conf. Comput. Vis. (ICCV), pp. 9992–10002, Oct. 2021, doi: 10.1109/ICCV48922.2021.00986.


\end{thebibliography}
\end{document}